\documentclass{article}
\usepackage{spconf,amsmath,graphicx}
\usepackage{booktabs}
\usepackage{tabularx}
\usepackage{graphicx}
\graphicspath{{Images/}}
\usepackage{hyperref}
\renewcommand{\paragraph}[1]{\noindent\textbf{#1}\quad}

\title{Improving Spoken Language Identification with Map-Mix}
\name{Shangeth Rajaa, Kriti Anandan, Swaraj Dalmia, Tarun Gupta, Eng Siong Chng}
\address{skit.ai, SCSE - NTU Singapore, \\ shangeth.rajaa@skit.ai }
\begin{document}
%
\maketitle
\begin{abstract}

\end{abstract}
The pre-trained multi-lingual XLSR model generalizes well for language identification after fine-tuning on unseen languages. However, the performance significantly degrades when the languages are not very distinct from each other, for example, in the case of dialects. Low resource dialect classification remains a challenging problem to solve. We present a new data augmentation method that leverages model training dynamics of individual data points to improve sampling for latent mixup. The method works well in low-resource settings where generalization is paramount. Our datamaps-based mixup technique, which we call Map-Mix improves weighted F1 scores by 2\% compared to the random mixup baseline and results in a significantly well-calibrated model. The code for our method is open sourced on \href{https://github.com/skit-ai/Map-Mix}{github}.


\begin{keywords}
data augmentation, mixup, datamaps, language identification, XLSR
\end{keywords}
\section{Introduction}
\label{sec:intro}


Spoken Language Identification(SLID) is the problem of classifying the language spoken by a speaker in an audio clip. SLID is useful in personalized voice assistants, automatic speech translation systems, multi-lingual speech recognition systems and has been used in call centers to route calls to a specific language operator automatically. Earlier studies \cite{tong2006integrating} \cite{ng2010prosodic} have used the phonetic, phonotactic, prosodic, and lexical features for SLID. Classical SLID models first extract the i-vectors \cite{dehak2011language} or x-vectors \cite{snyder2018spoken} and then train an independent classifier model on top. Acoustic features such as MFCC's and filter-banks are also commonly used as input features \cite{miao2019new}.

Recent advancement in deep learning has led to most works using Deep Neural Networks(DNN), Convolutional Neural networks(CNN) \cite{singh2021spoken} and Transformers \cite{tjandra2022improved} for SLID in an end-to-end manner. Although end-to-end models outperform classical methods on language id, they require a large amount of labeled training data, limiting their applicability to low-resource languages. Self Supervised Learning(SSL) based models \cite{fan2020exploring} and transfer learning can be used to aid the low-resourced data problem. 

Many data augmentation methods \cite{park2019specaugment} \cite{ko2017study} have been used to improve performance on speech tasks. These methods are used to increase the size of training data and help improve generalization. Recently, mixup based methods \cite{DBLP:journals/corr/abs-1710-09412} have been applied to multiple deep learning problems as a data augmentation technique to avoid over-fitting and have shown promising results in several audio-related tasks such as low-resourced speech recognition \cite{meng2021mixspeech} and environment sound classification \cite{zhang2018deep}. Few works have explored mixup techniques for the SLID task as a data augmentation method. 


\begin{figure}[h]
\centering
  \includegraphics[width=0.85\linewidth]{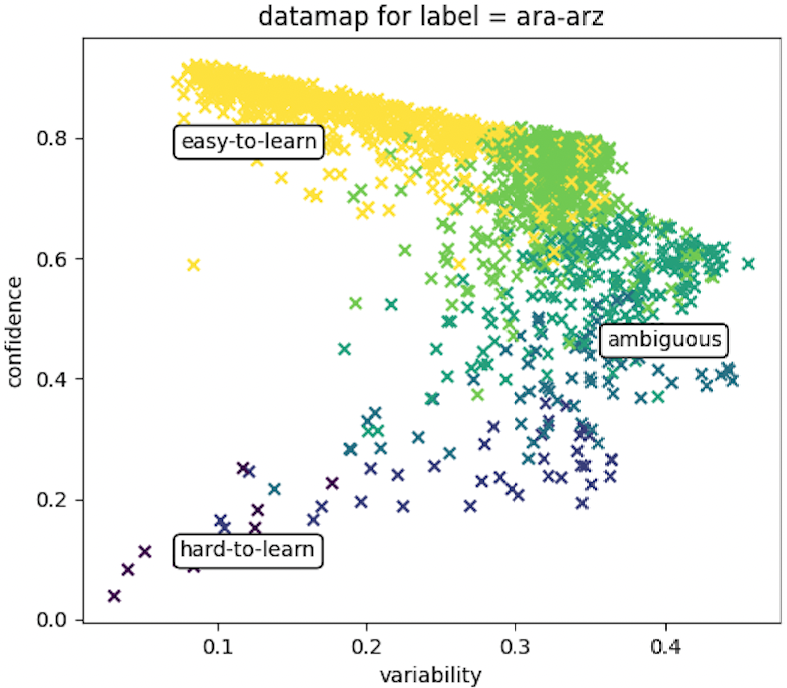}
  \caption{Datamaps for the dialect class "ara-arz" that is used for training our model. The training data is split into 3 different regions based on model training dynamics.}
  \label{fig:data_maps}
\end{figure}

\paragraph{Our Contribution:}We propose Map-Mix, a novel data augmentation method based on datamaps \cite{swayamdipta2020dataset}. Datamaps uses model training dynamics to split the dataset into three regions: easy-to-learn, hard-to-learn and ambiguous. Data points are mixed from different regions of the datamap to improve performance on low-resourced language ID. The use of probabilistic confidence labels instead of one-hot encoding further boosts performance. The XLSR model \cite{DBLP:journals/corr/abs-2111-09296} which is the SOTA model on VoxLingua107  is used to encode the speech signal. We compare our method with other mixup techniques and different settings of datamap based mixup. The only method that we've come across which is in similar veins to our is \cite{park2022calibration}. They use saliency maps to improve performance over an NLU task.



\section{Method}
\label{sec:method}

\subsection{Mixup}
\label{sec:mixup}
Mixup \cite{DBLP:journals/corr/abs-1710-09412} is a data augmentation techniques where synthetic data points are created $(\tilde{x}, \tilde{y})$ by interpolating two randomly sampled data points $(x_i, y_i)$ and $(x_j, y_j)$ from the training dataset $D$.

\begin{equation}
\label{eq:1}
\begin{aligned}
\tilde{x} = \lambda x_i + (1-\lambda) x_j \\
\tilde{y} = \lambda y_i + (1-\lambda) y_j
\end{aligned}
\end{equation}

The samples are mixed according to Equation \ref{eq:1}, where $\lambda \sim Beta(\alpha, \alpha)$ and $\alpha > 0$. Vanilla mixup is considered a static data augmentation technique since it can be performed before training begins. We refer to vanilla mixup as static mixup henceforth.

One limitation of static mixup is that it can only be applied to data points of the same dimensionality. Therefore, it cannot be used on raw textual or speech data with varying lengths. To overcome this limitation, Chen et al. \cite{chen2020mixtext} proposed a mixup variant called latent mixup, where the latent embeddings are mixed instead of the raw features. This method is dynamic since it requires a model to extract latent embeddings for interpolation.


\subsection{Datamaps}
Swayamdipta et. al. \cite{swayamdipta2020dataset} introduce Dataset Cartography as a tool to map and diagnose a dataset. Based on the behavior of individual data points during model training, we come up with a plot as shown in Figure \ref{fig:data_maps}. The confidence represents the model's prediction confidence for the true class and the variability is the model's variability in this confidence score.

Based on this map, all the data points are categorized into three distinct and disjoint regions: easy-to-learn, hard-to-learn and ambiguous. The authors conclude, easy-to-learn examples contribute to model convergence and faster optimization; the ambiguous data points help in generalization, and the hard-to-learn samples are likely annotation errors.

\subsection{Map-Mix}
In this paper, we combine datamaps with latent mixup. Instead of random sampling for the mixup, data points are sampled from specific regions obtained from datamaps. The datamaps are generated by fine-tuning an XLSR model.

The different variants experimented with are discussed in section \ref{expts:mapmix}. Our proposed method, Map-Mix, mixes data points belonging to the easy and ambiguous regions and removes the hard-to-learn samples from training. Further, instead of one-hot label encodings, we add confidence labels which are heuristically generated based on the ratio of the relative distribution of dialects in the language class.

\begin{figure}[h]
\centering
  \includegraphics[width=0.65\linewidth]{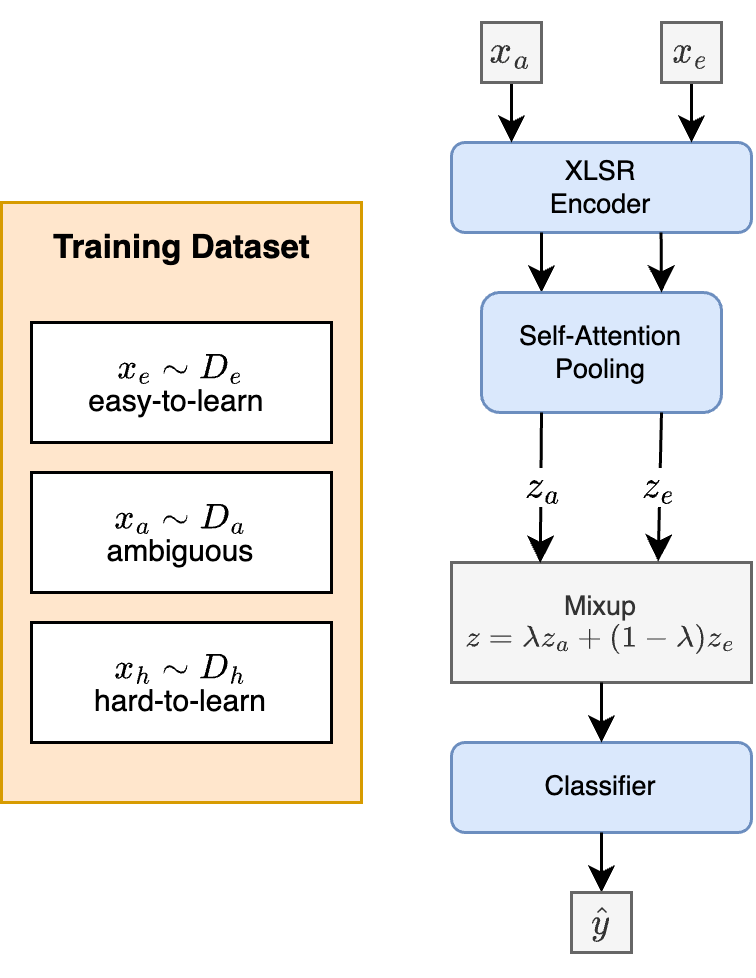}
  \caption{Model architecture and inputs used in Map-Mix, our proposed approach uses datamaps-based mixup to combine easy-to-learn samples with ambiguous samples.}
  \label{fig:model}
\end{figure}

\section{Experiments}
\label{sec:typestyle}

\subsection{Dataset}
The LRE 2017 dataset \cite{nistlre2017} is used for all experiments. It contains 14 dialects from 5 language clusters, namely Arabic, Chinese, English, Iberian, and Slavic, and is considered a difficult dataset to learn on. A visualization of the TSNE embeddings of the language clusters and dialects is shown in Figure \ref{fig:tnse5} and \ref{fig:tnse}. The original LRE 2017 dataset consists of around 2000 hours of the audio corpus in total.

Since the target is a low resource setting, the number of hrs for each dialect is constrained to 5 hrs. The resulting dataset is around 65 hours and corresponds to 3\% of the original dataset, and this is the subset that we present results on. To avoid sampling bias, three such subsets are constructed, each randomly sampled from the original dataset. The average metrics over the 3 subsets are reported for each model. The evaluation set used for the experiments is the same as the original dataset. 


\subsection{Pretrained Baselines}
Three pretrained Self Supervised Learning(SSL) models, wav2vec2 \cite{DBLP:journals/corr/abs-2006-11477}, Hubert \cite{DBLP:journals/corr/abs-2106-07447} and XLSR \cite{DBLP:journals/corr/abs-2111-09296} are used as baselines as they have shown to perform well for this task. As shown in Figure \ref{fig:model}, the encoder is followed by a self-attention pooling layer which reduces the temporal dimension of the encoded representations. These are then passed through a linear layer to predict the class label. The XLSR model is chosen as the encoder for all subsequent experiments as it performs the best amongst the pretrained baselines.

\subsection{Mixup Baselines}
The proposed method is compared against four mixup variants. Static and latent mixup are discussed in section \ref{sec:mixup}. For latent mixup(within), only data points belonging to dialects within the same language cluster are mixed. For latent mixup(across), data points belonging to dialects part of distinct language clusters are mixed.
   
\subsection{Map-Mix}
\label{expts:mapmix}
To improve over the baseline mixup variants, we experiment with 3 variants of datamaps-based mixup. For all the 3 approaches, the same model settings as the baseline mixup methods are used, the only difference being the sampling method for the latent mixup. 

\begin{itemize}
     \item Easy Mixup: Points from the entire training set are mixed with points only from the easy-to-learn region.
     \item Hard Mixup: Points from the entire training set are mixed with points only from the hard-to-learn region.
     \item Ambiguous + Easy Mixup: The points belonging to the hard-to-learn region are completely removed from the training set. The easy-to-learn samples and mixed with the ambiguous samples.
\end{itemize}


Our proposed method, which we call Map-Mix, is built on top of Ambiguous + Easy Mixup. The only difference is that the one-hot labels are replaced with confidence labels.

\subsection{Training Setup}
For fair evaluation, the training hyperparameters and settings are kept the same across all runs. The audio corpus is resampled to a sampling rate of 16kHz, to match the setting of the pretrained SSL baselines. For all the mixup experiments, $\alpha$ is set to 0.5, following \cite{DBLP:journals/corr/abs-1710-09412}.
For all the experiments, the complete model along with the pretrained encoder is finetuned using an Adam optimizer with a learning rate of $1e^{-5}$ for 50 epochs, and cross-entropy is used as the loss function.

\subsection{Evaluation Metrics}
For evaluation, each audio signal in the eval set is split into chunks of 8 secs with strides of 3 secs. The final prediction is done using the average softmax output of all the chunks.

Four metrics are reported for the 14 dialect classification problem. The classification accuracy(Acc), the weighted F1 score(WF1), the language cluster accuracy(C.Acc), and the expected calibration error(ECE).

\section{Results and Discussion}
\label{sec:majhead}

\paragraph{SSL and Mixup Baselines}
In Table \ref{tab:result-table} amongst the 3 pretrained baselines, the XLSR model gives the best WF1 score. This is expected as XLSR is a multi-lingual model whereas Hubert and Wav2vec2.0 are monolingual English models. We choose the XLSR model as the base encoder model for all subsequent experiments. 

\begin{figure}[]
\centering
  \includegraphics[width=0.75\linewidth]{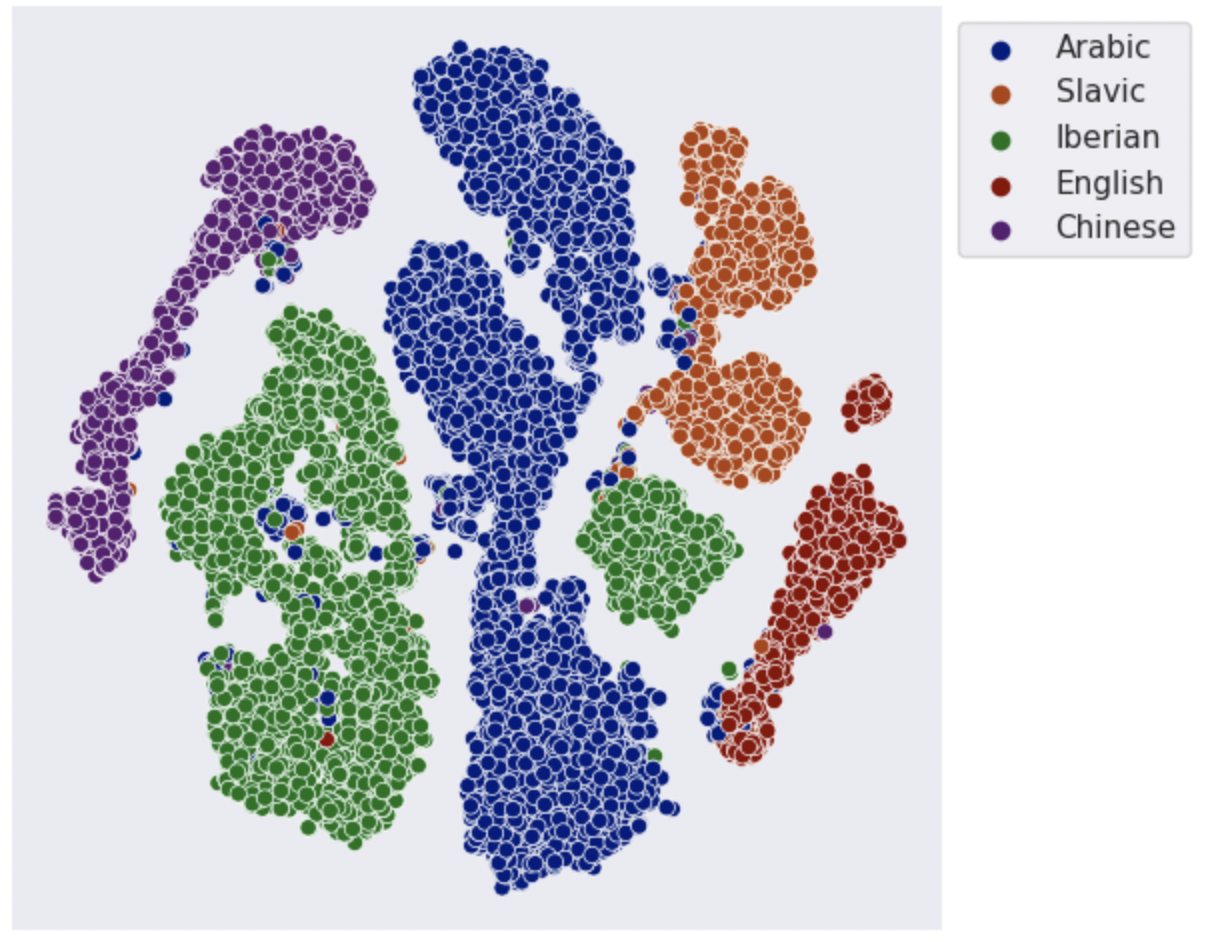}
  \caption{TSNE plot of the XLSR embeddings trained using Map-Mix for the 5 language clusters in the eval set of LRE-17. }
  \label{fig:tnse5}
\end{figure}

Latent mixup outperforms static mixup by around 11\% WF1 and has faster convergence. Mixing data points in the embedding space proves better for generalization. We compare the different sampling methods for the latent mixup in Table \ref{tab:result-table}. As seen in Figure \ref{fig:tnse5}, the 5 major language clusters are well separated. This can also be inferred from the high C.Acc obtained. Mixing across clusters performs better than mixing within clusters in this setting as mixing across clusters reduces data sparsity between clusters and empirically has better generalization on the eval set. 


\begin{table}[h]
\centering
\begin{tabular}{@{}lrrrr@{}}
\toprule
\textbf{Model} & \multicolumn{1}{l}{\textbf{Acc}} & \multicolumn{1}{l}{\textbf{WF1}} & \multicolumn{1}{l}{\textbf{C.Acc}} & \multicolumn{1}{l}{\textbf{ECE}} \\ \midrule
wav2vec & 0.522 & 0.481 & 0.904 & 0.271 \\
hubert & 0.541 & 0.505 & 0.9 & 0.231 \\
XLSR & 0.632 & 0.601 & \textbf{0.959} & 0.158 \\ \midrule
static mixup & 0.564 & 0.529 & 0.842 & 0.112 \\
latent mixup(random) & 0.658 & 0.638 & 0.945 & 0.108 \\
latent mixup(within) & 0.646 & 0.614 & 0.945 & 0.127 \\
latent mixup(across) & 0.655 & 0.641 & 0.949 & 0.105 \\ \midrule
Map-Mix(proposed) & \textbf{0.677} & \textbf{0.658} & 0.958 & \textbf{0.067} \\ \bottomrule
\end{tabular}
\caption{Results for the SSL baselines, the mixup baselines, and the Map-Mix method on the LRE-2017 evaluation set.}
\label{tab:result-table}
\end{table}

\begin{table}[h]
\centering
\begin{tabular}{llrrr}
\hline
\textbf{Model} & \textbf{Acc} & \multicolumn{1}{l}{\textbf{WF1}} & \multicolumn{1}{l}{\textbf{C.Acc}} & \multicolumn{1}{l}{\textbf{ECE}} \\ \hline
easy mixup & 0.657 & 0.631 & 0.948 & 0.107 \\
hard mixup & 0.636 & 0.605 & 0.943 & 0.132 \\
amb + easy mixup & 0.658 & 0.644 & 0.945 & 0.101 \\ \hline
Map-Mix & \textbf{0.677} & \textbf{0.658} & \textbf{0.958} & \textbf{0.067} \\ \hline
\end{tabular}
\caption{Results for different datamaps-based mixup methods on the LRE-2017 evaluation set.}
\label{tab:datamap-mix-result}
\end{table}

\begin{figure}[]
\centering
  \includegraphics[width=0.8\linewidth]{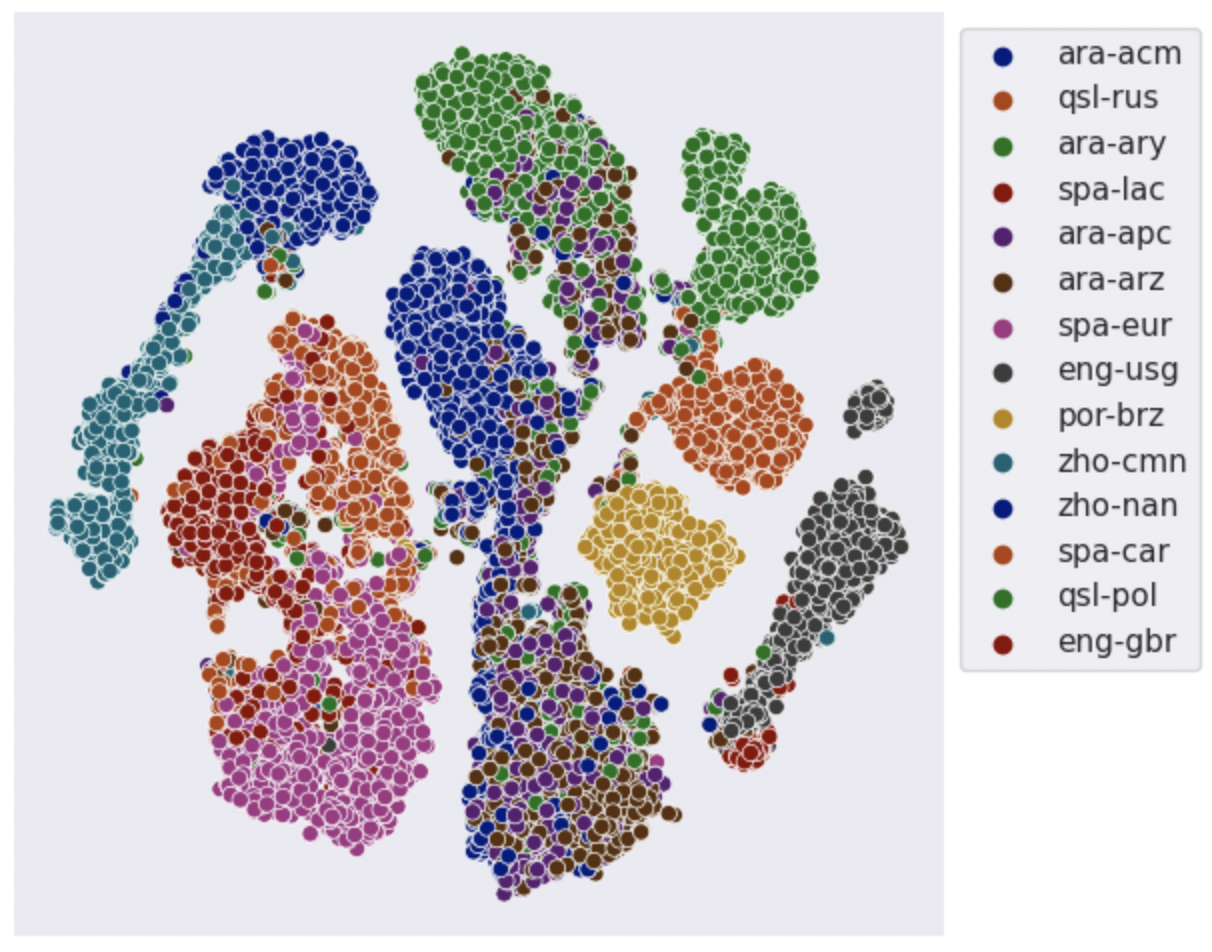}
  \caption{TSNE plot of the XLSR embeddings trained using Map-Mix method on the evaluation set of LRE-17. This figure shows the 14 dialects part of the 5 language clusters.}
  \label{fig:tnse}
\end{figure}

\hfill

\paragraph{Datamaps based Mixup}
In Table \ref{tab:datamap-mix-result}, we show the evaluation metrics for the datamaps based mixup methods. Easy mixup performs better than hard mixup. This is in line with prior findings \cite{swayamdipta2020dataset}. Easy samples help in better optimization, and the hard samples can be considered labeling errors. Based on this result, we remove the hard samples from the next experiment. The model trained with amb+easy mixup has a better WF1 score compared to easy mixup at the cost of a slight decrease in cluster accuracy. This is consistent with \cite{swayamdipta2020dataset}, as training the model on the ambiguous samples helps improve generalization and betters Out-of-Distribution(OOD) performance. 

\hfill

\begin{figure}[hbt!]
  \includegraphics[width=0.9\linewidth]{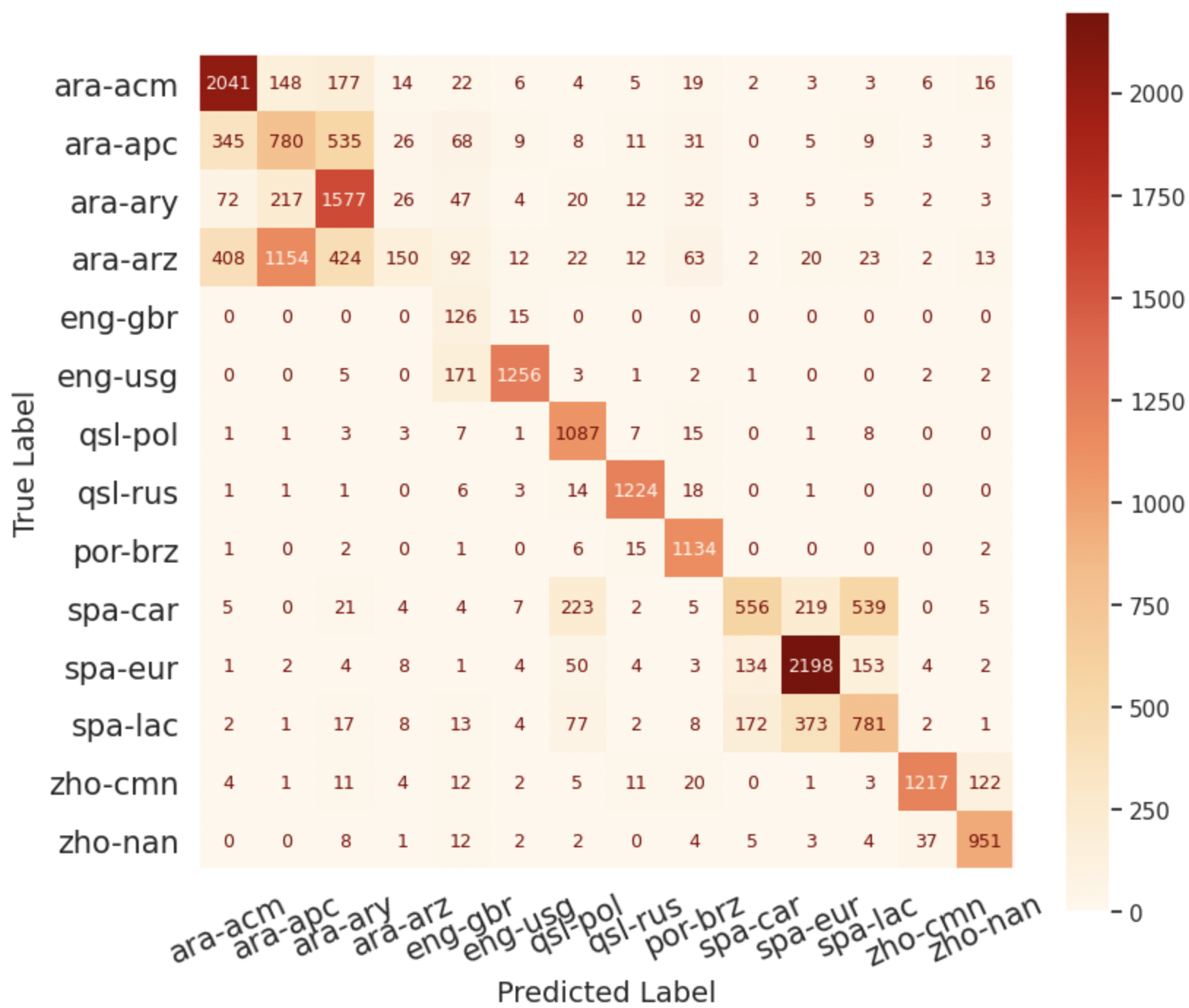}
  \caption{Confusion Matrix of the Map-Mix method on the evaluation set of LRE-17 dataset.}
  \label{fig:conf_matrix}
\end{figure}

\paragraph{Map-Mix}
Our proposed approach, which adds confidence labels to amb + easy mixup gives the best result for all metrics compared to the other methods, except cluster accuracy. For cluster accuracy, the XLSR baseline performs the best. The multi-lingual XLSR model does well for the 5 language clusters but has a low WF1 and is unable to generalize well to 14 class dialect classifications where the dialects are not very well separated(see Figure \ref{fig:tnse}). Map-Mix is able to nearly match the cluster accuracy of the XLSR baseline and improves total accuracy by 4.5\% and WF1 by 5.7\% over XLSR for the 14 dialect classification task. It improves the WF1 by 2\% compared to the static mixup baseline. The confusion matrix for Map-Mix is shown in Figure \ref{fig:conf_matrix}. The model's errors are due to misclassification within the language cluster, as can be seen by the presence of square weights along the diagonal. The Spanish and Arabic dialects are the most difficult to separate.

\hfill

\paragraph{Expected Calibration Error}
A well-calibrated DNN model produces confidence closely approximated by expected accuracy. DNNs trained with mixup are better-calibrated \cite{thulasidasan2019mixup}. From Table \ref{tab:result-table}, we can observe, all the mixup methods have better ECE than the baselines. The Map-Mix method further reduces the ECE and results in the best-calibrated model. Label smoothing \cite{muller2019does} and its variants help in better calibration of neural models. Confidence labels can be considered as a form of label smoothing with class similarity and therefore aid in reducing the ECE.



\section{Conclusion}
We proposed a new data augmentation method, Map-Mix which is based on mixup across different datamaps-based regions for the spoken language identification task. We benchmark this method on a low resource subset of a difficult dataset, the LRE 2017. This method performs well in a low-resource setting by identifying and removing samples from the training data which don't help in generalization. Map-mix provides faster model convergence and results in a better-calibrated model by improving on random sampling for mixup with datamaps.

\bibliographystyle{IEEEbib}
\bibliography{refs}

\end{document}